# LoD Sketch Extraction from Architectural Models Using Generative AI

*Dataset Construction for Multi-Level Architectural Design Generation*

Xusheng Du[1], Athiwat Kongkaeo[2], Ye Zhang[3], and Haoran Xie[4,5]

[1,4] *Japan Advanced Institute of Science and Technology.*
[2] *Chulalongkorn University.*
[3] *Tianjin University.*
[5] *Waseda University.*
[1] *s2320034@jaist.ac.jp, 0009-0008-1086-2081*

**Abstract.** For architectural design, the representation across multiple Levels of Details (LoD) is essential for achieving a smooth transition from conceptual massing to detailed modeling. However, traditional LoD modeling processes rely on manual operations, which are time-consuming, labor-intensive, and prone to geometric inconsistencies. The rapid advancement of generative artificial intelligence (AI) has opened new possibilities for generating multi-level architectural models from sketch inputs. Nevertheless, its application remains limited by the lack of high-quality paired LoD training data. To address this issue, we propose an automatic LoD sketch extraction framework using generative AI models, which progressively simplifies high-detail architectural models to automatically generate geometrically consistent and hierarchically coherent multi-LoD representation. The proposed framework integrates computer vision techniques with generative AI methods to establish a progressive extraction pipeline that transitions from detailed representations to volumetric abstractions. Experimental results demonstrate that the proposed method maintains strong geometric consistency across LoD levels, achieving SSIM values of 0.7319 and 0.7532 for the transitions from LoD3 to LoD2 and from LoD2 to LoD1, respectively. The corresponding normalized Hausdorff distances are 25.1% and 61.0% of the image diagonal, reflecting controlled geometric deviation during abstraction. These results verified that the proposed framework effectively preserves global structure while achieving progressive semantic simplification with different LoD, providing reliable data and technical support for AI-driven multi-level architectural generation and hierarchical modeling.

## 1. Introduction

In architectural design practice, architectural models are typically represented at different Levels of Detail (LoD) to accommodate the progressive transition from conceptual expression to detailed design. According to the CityGML standard (OGC, 2012), LoD1 denotes coarse volumetric blocks, LoD2 adds simplified roof geometry and semantic components, and LoD3 includes detailed façade elements such as windows and doors. However, constructing and maintaining these multi-LoD models in digital twin and Building Information Modeling (BIM) systems still rely heavily on manual reconstruction and refinement, which are time-consuming, labor-intensive, and prone to geometric inconsistencies across LoD levels.

The rapid advancement of Generative Artificial Intelligence (AI) has opened up new possibilities for architectural design. Generative models such as Stable Diffusion (Rombach et al., 2022) and conditional generation frameworks like ControlNet (Zhang et al., 2023) can synthesize architectural images from simple inputs such as text or sketches. This capability enables multi-level architectural representations across LoD stages based on sketch inputs, offering a novel paradigm for automated design exploration. Despite its enormous potential, the application of generative AI in architectural design still faces several critical challenges. First, the effective training of such models requires high-quality paired data across multiple LoD levels, yet existing public architectural datasets typically contain only a single LoD and lack paired samples that capture the evolution of geometric detail. Second, the extraction of LoD sketches often fails to ensure geometric consistency across levels and viewpoints, which limits the model's ability to learn reliable mappings. Third, the semantic ambiguity of sketch elements, such as the mixture of decorative textures and structural features, further complicates the learning. Therefore, automatically generating geometrically consistent multi-LoD sketch sequences from high-detail architectural models remains a key challenge for AI-driven architectural generation.

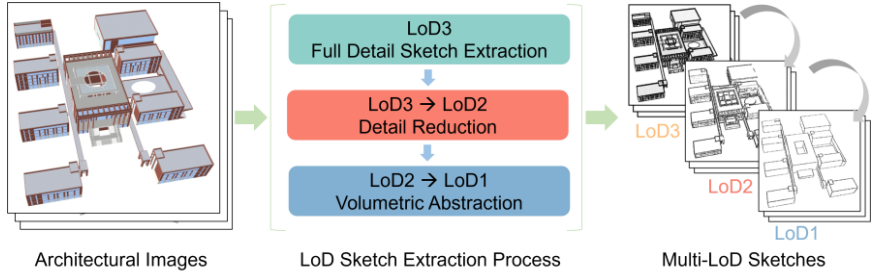

*Figure 1. Overview of the proposed Automatic LoD Sketch Extraction Framework.*

To address these challenges, we propose an automatic LoD sketch extraction method using generative AI models, aiming to automatically produce multi-level, geometrically consistent LoD sketches from high-detail architectural models (as shown in Figure 1). The proposed method integrates traditional computer vision techniques with generative AI models. Through the progressive process, we construct a multi-LoD sketch dataset that is structurally aligned and detail controllable. The main contributions of this work include: (1) an automated LoD sketch extraction framework that achieves geometrically consistent transformation from high-detail models to multi-level sketches; and (2) a paired multi-LoD sketch dataset, offering standardized training samples for generative AI-based multi-stage architectural modeling. In this context,

LOD SKETCH EXTRACTION FROM
ARCHITECTURAL MODELS USING GENERATIVE AI

LoD reduction serves as a foundational representation for AI-supported architectural design, enabling multi-level consistency modeling and providing structured training data to support subsequent detail refinement and design exploration.

## 2. Related Work

### 2.1. LOD MODELING

In Building Information Modeling (BIM) and urban digital twin systems, the Level of Detail (LoD) is a key concept for describing the geometric and semantic precision of architectural models (Leite et al., 2011; Biljecki et al., 2016). Previous studies have defined standardized hierarchies from LoD1 (massing and spatial layout) to LoD3 (detailed representation), providing a common framework for multi-stage design. However, most existing automated modeling approaches focus on refining high-LoD models, while the reverse simplification—from LoD3 to lower levels—has received limited attention. The lack of multi-LoD datasets with clear geometric alignment and inter-level mapping has further restricted progress in automated simplification and hierarchical model control (Kutzner et al., 2020).

### 2.2. SKETCH EXTRACTION

Sketches serve as an essential medium in architectural design, bridging conceptual thinking and spatial representation. Automatic sketch extraction and structural interpretation have been long-standing topics in computer vision and digital architecture. Classical methods such as Canny, Sobel, and Laplacian filters can extract contours and line features from renderings or photographs but are often sensitive to illumination and texture noise. Recent deep learning approaches, including Holistically-Nested Edge Detection (HED) and Richer Convolutional Features (RCF), enable multi-scale and semantically consistent line extraction (Xie & Tu, 2015; Liu et al., 2019). Furthermore, learning-based sketch extraction methods have been developed to generate structured line drawings of architectural and heritage images with improved geometric consistency (Dong, 2025). Nevertheless, achieving consistent abstraction of structural semantics across varying levels of detail remains a key challenge for automated sketch-based architectural modeling.

### 2.3. GENERATIVE AI FOR ARCHITECTURAL DESIGN

In recent years, generative models such as Stable Diffusion (Rombach et al., 2022), ControlNet (Zhang et al., 2023), and T2I-Adapter (Mou et al., 2024) have demonstrated the ability to synthesize high-fidelity and semantically coherent architectural images from multimodal inputs, including text, sketches, and depth maps. These techniques have significantly advanced research on sketch-to-façade generation and multi-stage design representation across different LoD levels (Li et al., 2024; Zhang et al., 2024; Pan et al., 2025). However, the effectiveness of such models largely depends on the availability of high-quality paired datasets. The architectural domain still lacks multi-LoD datasets with precise geometric alignment and well-defined correspondences between levels, which limits progress in learning structural abstraction and controllable generation. To address this issue, we propose an automatic

LoD sketch extraction method that integrates computer vision and generative modeling to map high-LoD models to lower-LoD sketches, thereby establishing a data foundation for future AI-based architectural generation.

## 3. Method

### 3.1. FULL-DETAIL SKETCH EXTRACTION

At the first stage (LoD3), the objective is to extract full-detail architectural sketches that faithfully preserve all visible geometric edges from rendered images while explicitly suppressing non-structural visual factors. This stage focuses on retaining both major structural contours and fine architectural details, such as window frames, façade edges, and texture details, while removing shadows, illumination variations, and rendering noise that are irrelevant to geometric representation. The resulting LoD3 sketches serve as a geometry-complete reference for subsequent abstraction stages.

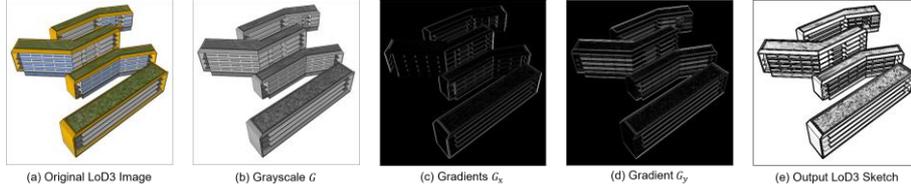

*Figure 2. Intermediate results of the full-detail sketch extraction process at the LoD3 stage.*

To achieve this, we design a full-detail sketch extraction pipeline based on image gradient analysis and morphological enhancement. Figure 2 shows the intermediate results of the full-detail sketch extraction process at the LoD3 stage. First, the input color rendering, obtained from a 3D architectural model, is converted into a grayscale image $G$, and the Sobel operator is applied to compute horizontal and vertical gradient components $G_x$ and $G_y$, respectively. The gradient magnitude M is then calculated as:

$$M = \sqrt{G_x^2 + G_y^2},$$

which is followed by an inversion operation $(255 - M)$ to obtain an edge brightness distribution consistent with human visual perception, resulting in an edge map $E$. Next, we introduce two adjustable parameters, the shadow blending coefficient $\alpha$ and the line thickness coefficient $\beta$, to control the retention of grayscale information and the enhancement of line intensity, respectively. The overall transformation is defined as:

$$S = (1 - \alpha)E + \alpha G,$$

$$O = (S - 128) \times (1 + 2\beta) + 128,$$

where $S$ represents the intermediate blended result and $O$ denotes the final sketch output. The constant 128 serves as the mid-gray reference point, ensuring that global brightness remains stable during contrast enhancement. A smaller $\alpha$ produces a purer line-drawing appearance, while a larger $\alpha$ preserves more shading information. $\beta$ controls the strength and thickness of the lines. Higher values produce darker and bolder contours, whereas lower values yield finer and more delicate strokes.

Finally, we apply morphological Black-Hat enhancement to darken fine-line regions, improving line contrast and visual uniformity without affecting broader grayscale areas. This process ensures that the generated LoD3 sketches maintain both structural fidelity and visual clarity, providing a reliable foundation for subsequent LoD simplification stages.

## 3.2. DETAIL REDUCTION USING GENERATIVE MODELING

The second stage (from LoD3 to LoD2) aims to reduce components and texture details in architectural models, enabling a transition from detailed representations to simplified structural forms. However, directly training a model using paired LoD3 and LoD2 sketches fails to produce meaningful results. After sketch extraction, the two levels exhibit similar edge distributions, where texture and structural lines become intertwined, leaving subtle geometric differences. As a result, the model struggles to discern which lines should be preserved or removed, leading to unstable convergence and inconsistent detail reduction. Moreover, sketches inherently contain only edge information while lacking mid-level semantic cues such as lighting, materials, and shading, which are factors essential for understanding the concept of detailed level.

To address these challenges, we adopt a two-stage approach. Specifically, we first convert LoD3 images into LoD2-style RGB images, allowing the model to learn the "fine-to-coarse" transformation in the image domain, where richer visual features are available. In the second stage, sketches are extracted from the generated LoD2 images, producing structurally clear and detail-simplified LoD2 sketches. this indirect image-based approach provides stronger learning signals and better geometric consistency in the generated results.

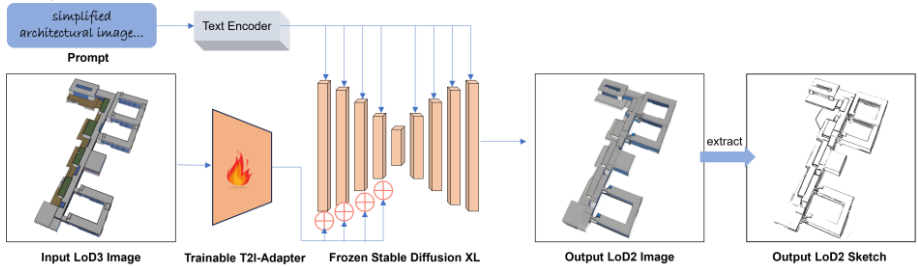

*Figure 3. Detail reduction from LoD3 to LoD2.*

As illustrated in Figure 3, during the first stage we employ the Stable Diffusion XL (Podell et al., 2023) architecture integrated with a T2I-Adapter (Mou et al., 2024) module to generate LoD2-style images from LoD3 inputs. We adopt the T2I-Adapter instead of the ControlNet (Zhang et al., 2023) architecture because its mechanism of feature modulation and injection is better suited for flexible detail abstraction. ControlNet imposes strong spatial constraints on the generation process and often forces the model to strictly follow the fine-grained edge structures of the LoD3 input. In contrast, the T2I-Adapter provides a lighter and more adaptive conditioning pathway that allows the model to learn a soft mapping from complex to simplified visual features. This flexibility enables the network to effectively suppress façade textures and window details while preserving global proportions and spatial coherence.

During training, the LoD3 image serves as the conditional input, and the corresponding LoD2 image is used as the generation target. Short textual prompts (e.g., "LoD2 style, simplified structure, no small fixtures or textures") are provided for lightweight semantic guidance. Through this process, the model learns to perform a visual mapping from complex to simplified structural representations, generating RGB images that exhibit LoD2-level abstraction. Next, we transform the generated LoD2 image into its sketch representation using the same extraction process described in Section 3.1. The generated LoD2 sketches retain consistent building mass boundaries, roof outlines, and floor separations, while effectively removing fine-grained details such as window frames, surface textures, and decorative patterns.

### 3.3. VOLUMETRIC ABSTRACTION WITH COTROLNET

The third stage (from LoD2 to LoD1) aims to further abstract the LoD2 sketches into LoD1 representations that preserve only the primary building volumes. Since LoD2 and LoD1 differ significantly in both semantic hierarchy and geometric structure, using a single conditional input often fails to achieve a balance between shape preservation and spatial abstraction. To address this issue, we design a dual-ControlNet abstraction framework, which jointly leverages structural sketches and depth information to learn the mapping from detailed structures to simplified volumetric forms.

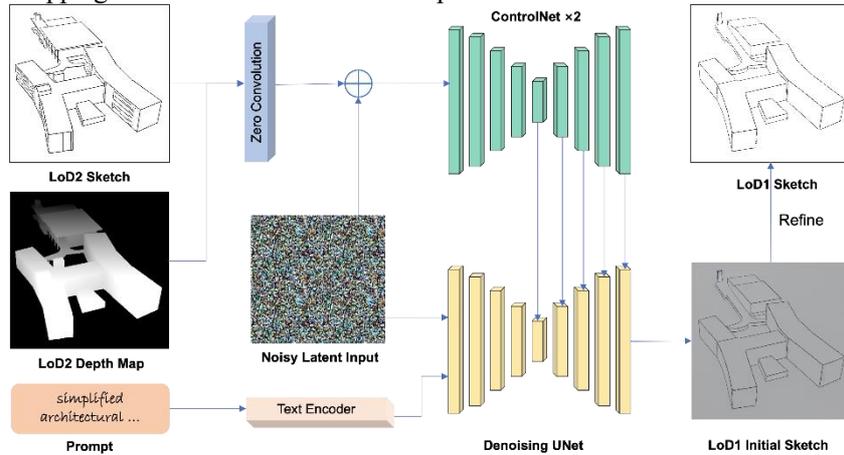

*Figure 4. Volumetric abstraction from LoD2 to LoD1 using a dual-ControlNet framework.*

During training, the model takes the LoD2 sketch and its corresponding depth map as dual conditioning inputs, which are processed by two specialized branches. The Sketch ControlNet preserves the global contour and geometric proportions of the building, preventing structural distortion or misalignment during abstraction. The Depth ControlNet provides constraints on spatial hierarchy and volumetric relationships, enabling the model to remove small components such as doors and windows while maintaining correct depth ordering between building masses.

The multi-scale features extracted from both Sketch and Depth ControlNets are injected into the main UNet backbone at down sampling and intermediate layers, where they modulate the latent diffusion process with the noise embeddings. This mechanism allows the model to achieve a progressive transition from detailed structures to abstract volumes during denoising. Training is guided by the standard noise-consistency loss of diffusion, minimizing the difference between the predicted and true noise:

$$L_{\text{train}} = \mathbb{E}_{t,\epsilon} \parallel \epsilon_\theta(x_t, t, c_{\text{sketch}}, c_{\text{depth}}) - \epsilon \parallel^2,$$

where $c_{\text{sketch}}$ and $c_{\text{depth}}$ denote the sketch and depth conditioning inputs, respectively. The model is trained end-to-end on LoD2–LoD1 pairs to automatically remove small-scale features such as openings, ornaments, and façade details while preserving structural integrity.

During inference, the system requires a LoD2 sketch to generate the LoD1 abstraction. For scenes without depth annotations, we employ the MiDaS (Ranftl et al., 2020) pretrained model to estimate a depth map from the input LoD2 sketch. This estimated depth is then fed into the Depth ControlNet as auxiliary guidance, together with the sketch condition, to support volumetric abstraction generation. Throughout the multi-step diffusion sampling process, the model progressively removes fine details while retaining the main building volumes and spatial hierarchy. Finally, the generated LoD1 image is post-processed using Canny edge detection and line refinement to obtain a geometrically consistent and hierarchically clear LoD1 sketch.

## 4. Experiments and Results

### 4.1. EXPERIMENTAL SETUP AND DATA PREPARATION

A total of 50 groups of architectural models were constructed including LoD1, LoD2, and LoD3 representations of 150 models. Each model was rendered using Blender and a pyrender-based orbit-capture script with 36 azimuth angles (0°–350°, step 10°) and 7 elevation angles (0°–60°), producing 252 viewpoints per LoD. For each viewpoint, both RGB and depth images were rendered at a resolution of 512 × 512 pixels, resulting in 504 images per model. Across all LoD levels and 50 model groups, the dataset contained approximately 75,600 images (37,800 RGB and 37,800 depth). All data were standardized and geometrically aligned to maintain consistency across LoD levels. This dataset provides comprehensive paired samples for the multi-stage generation framework, which supports the experiments on detail extraction, structural reduction, and volumetric abstraction. All experiments were conducted on a workstation equipped with a GeForce RTX 5090 GPU. The experimental framework consists of three main stages, corresponding to different LoD transitions. All experiments were performed under fixed random seeds and unified rendering settings to ensure reproducibility.

### 4.2. QUALITATIVE RESULTS

The proposed framework demonstrates strong capability in generating hierarchical sketch representations that progressively abstract architectural structures across LoD levels. As illustrated in Figure 5, the process begins with previously unseen LoD3 images, from which LoD3 sketches are extracted to capture the primary structural lines and compositional organization of the buildings. Subsequently, through the generative transformation from LoD3 to LoD2, the model removes secondary textures and decorative details while preserving the essential geometric boundaries, resulting in sketches that emphasize clean structural outlines and spatial coherence. Finally, the LoD1 sketches produced by the dual-ControlNet abstraction stage depict simplified, shoebox-like volumes that express the core building masses and their spatial hierarchy. Across all levels, the generated sketches exhibit high geometric consistency and accurate proportion alignment, ensuring smooth transitions between abstraction stages.

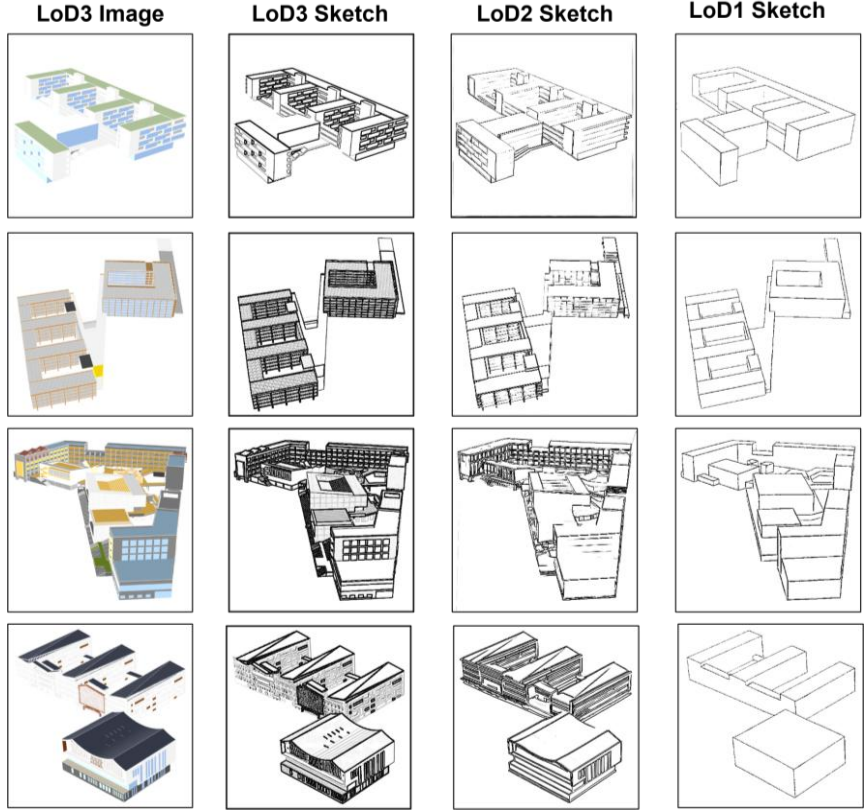

*Figure 5. Qualitative results showing the progressive abstraction from LoD3 image to LoD1 sketch.*

### 4.3. QUANTITATIVE EVALUATION

The quantitative comparison shows that the metric distributions of the two stages align precisely with their respective functional objectives, confirming the rationality and effectiveness of the proposed hierarchical simplification framework. All images are standardized to a resolution of 512×512, corresponding to a diagonal length of approximately 724.1 px, which allows the Hausdorff Distance (HD) to be normalized for scale consistency across stages.

As summarized in Table 1, during the LoD3-to-LoD2 stage, the model achieves an SSIM of 0.7319, an MSE of $5.15\times10^3$, and a normalized Hausdorff Distance (HD) equal to 25.1% of the image diagonal. These results demonstrate that the proposed generative simplification effectively eliminates local textures and redundant line segments while preserving the overall geometric configuration of the building. The MSE accounts for only 7.9% of the maximum possible pixel error (65,025 for 8-bit images), indicating that most changes occur at small-scale structural features rather than in the global geometry. This trend substantiates the intended role of this stage—detail reduction while maintaining geometric fidelity.

In contrast, during the LoD2-to-LoD1 stage, the SSIM slightly increases to 0.7532, and the MSE further decreases to 4240 (approximately 6.5% of the 8-bit range), reflecting globally cleaner and more homogeneous sketch representations. However, the HD rises sharply to 441.65 pixels (61.0% of the image diagonal). This increase should not be interpreted as degradation; rather, it reflects the intentional geometric abstraction inherent to volumetric simplification, where the model deliberately removes façade elements—such as windows, decorative features, and fine edges—and reconstructs the outline into simplified massing boxes. The higher HD quantitatively encodes the semantic transition from structural sketches to volumetric representations.

Table 1. Quantitative comparison of two LoD transition stages.

| Stage | SSIM ↑ | MSE ↓ | HD (px) ↓ | Normalized HD (% diag.) ↓ |
|---|---|---|---|---|
| LoD3 → LoD2 | 0.7319 | $5.15 \times 10^3$ | 181.90 | 25.1% |
| LoD2 → LoD1 | 0.7532 | $4.24 \times 10^3$ | 441.65 | 61.0% |

## 5. Conclusion

In this work, we proposed an LoD sketch extraction framework to automatically generate geometrically consistent and hierarchically coherent multi-LoD sketches from high-detail architectural models. Through three key stages: (1) full-detail sketch extraction, (2) generative detail reduction, and (3) volumetric abstraction, the framework achieves continuous simplification from LoD3 to LoD1 while maintaining structural proportions and spatial alignment. Experiments demonstrate that the proposed pipeline effectively removes redundant details, preserves geometric integrity, and produces visually coherent LoD sequences across various architectural types. Quantitative metrics (SSIM, MSE, and HD) confirm the balance between fidelity and abstraction at each level. The LoD3-to-LoD2 transition relies on an intermediate image generation step, and variations in the generated image quality may occasionally lead to locally ambiguous or inconsistent LoD2 sketch semantics. The constructed multi-LoD sketch dataset provides a standardized foundation for AI-driven architectural generation. Future work will focus on broader benchmarking against alternative LoD abstraction methods, expanding dataset diversity, conducting perceptual evaluations, and performing framework-level analyses of design choices, parameter sensitivity, and auxiliary input reliability to further enhance robustness. In addition, we aim to integrate LoD sketches with BIM semantics toward an automated sketch-to-BIM pipeline.

## Acknowledgements

This work was supported by JST SPRING, Japan Grant Number JPMJSP210, JST BOOST Program Japan Grant Number JPMJBY24D6, and the National Natural Science Foundation of China, Grant Number 52508023.

**Attribution**

ChatGPT (OpenAI, 2025) was used to translate the manuscript and to improve flow.